\documentclass[conference]{IEEEtran}
\IEEEoverridecommandlockouts
\usepackage{cite}
\usepackage{amsmath,amssymb,amsfonts}
\usepackage{hyperref}
\usepackage{tikz}
\ifCLASSOPTIONcompsoc
    \usepackage[caption=false, font=normalsize, labelfont=sf, textfont=sf]{subfig}
\else
\usepackage[caption=false, font=footnotesize]{subfig}

\usepackage{algorithm}
\usepackage[]{algpseudocode}

\usepackage{tabularx}

\newcommand\clearrow{\global\let\rowmac\relax}
\clearrow
\usepackage{graphicx}
\usepackage{textcomp}
\usepackage{xcolor}

\DeclareMathOperator*{\argmin}{arg\,min}
\def\BibTeX{{\rm B\kern-.05em{\sc i\kern-.025em b}\kern-.08em
    T\kern-.1667em\lower.7ex\hbox{E}\kern-.125emX}}
\begin{document}


\title{Generalizable Cross-Graph Embedding for GNN-based Congestion Prediction\\}

\author{\IEEEauthorblockN{Amur Ghose \IEEEauthorrefmark{1} \IEEEauthorrefmark{2}, Vincent Zhang \IEEEauthorrefmark{1} \IEEEauthorrefmark{2}, Yingxue Zhang \IEEEauthorrefmark{1} \IEEEauthorrefmark{2}, Dong Li \IEEEauthorrefmark{2}, Wulong Liu \IEEEauthorrefmark{2}, Mark Coates \IEEEauthorrefmark{3}}}

\maketitle

\begin{abstract}
Presently with technology node scaling, an accurate prediction model at early design stages can significantly reduce the design cycle. Especially during logic synthesis, predicting cell congestion due to improper logic combination can reduce the burden of subsequent physical implementations. There have been attempts using Graph Neural Network (GNN) techniques to tackle congestion prediction during the logic synthesis stage. However, they require informative cell features to achieve reasonable performance since the core idea of GNNs is built on the message passing framework, which would be impractical at the early logic synthesis stage. To address this limitation, we propose a framework that can directly learn embeddings for the given netlist to enhance the quality of our node features. Popular random-walk based embedding methods such as Node2vec, LINE, and DeepWalk suffer from the issue of cross-graph alignment and poor generalization to unseen netlist graphs, yielding inferior performance and costing significant runtime. In our framework, we introduce a superior alternative to obtain node embeddings that can generalize across netlist graphs using matrix factorization methods. We propose an efficient mini-batch training method at the sub-graph level that can guarantee parallel training and satisfy the memory restriction for large-scale netlists. We present results utilizing open-source EDA tools such as DREAMPLACE and OPENROAD frameworks on a variety of openly available circuits. By combining the learned embedding on top of the netlist with the GNNs, our method improves prediction performance, generalizes to new circuit lines, and is efficient in training, potentially saving over $90 \%$ of runtime.

\end{abstract}

\section{Introduction}
Routability-aware or congestion-aware placement has been well-studied in the existing literature \cite{lin2014polar, li2007routability, xie2018routenet}. By leveraging the rough evaluation of congestion via RUDY \cite{spindler2007fast} and NCTU-GR~\cite{dai2011nctu, liu2013nctu}, the candidate placement can be optimized iteratively. However, with the technology node scaling and the increase of chip size, each placement iteration could be very time-consuming in order to achieve the optimal placement objective with good routability or less congestion. In addition, it would remain difficult to alleviate the routing congestion even with unlimited iterations to implement the best placement results, which is actually induced by the improper logic implementation at the early design stage, e.g., using too many high fan-in and fan-out logic cells in the logic synthesis phase.   \par

\begingroup\renewcommand\thefootnote{\IEEEauthorrefmark{2}}
\footnotetext{Noah's Ark Lab, Huawei \IEEEauthorrefmark{3} McGill University}
\endgroup

\begingroup\renewcommand\thefootnote{}
\footnotetext{\IEEEauthorrefmark{1} \{amur.ghose, vincent.zhang2, yingxue.zhang \}@huawei.com (Corresponding authors) }
\endgroup

This naturally raises the problem of estimating logic-induced congestion from the netlist directly, prior to the placement iterations. This approach is explored in a recent work, CongestionNet \cite{kirby2019congestionnet}. Using a deep Graph Attention Network (GAT) \cite{velivckovic2017graph}, CongestionNet achieves relatively promising congestion prediction results in the logic synthesis phase. As the core idea of GNNs is built on top of the message passing framework, the work assumes that informative cell features can be used to achieve reasonable performance. However, this would be impractical at the early logic synthesis stage. \par

In addition, the aforementioned GAT is a subclass of the broader group of Graph Neural Networks (GNN) \cite{kipf2016semi, hamilton2017}. Such GNN methods are effective for general prediction tasks on graph structured data. In the case of circuits, the key deviation from the usual use case of GNNs is that the train and test graphs can be completely disjoint. Given some congestion data for a known set of circuits, the desired prediction is often {not} on the unseen cells of the same circuits, but on completely new circuits. This more challenging case is called \textbf{inductive learning}, and it requires specific GNN design and training approaches. \par

Separate from learning-based methods, there exist methods like GTL \cite{jindal2010detecting} which estimate congestion directly from the netlist based on the graph structure. Analogously, in the key areas of GNN usage such as recommendation systems and node level regression, there exist methods for learning high quality \textbf{embedding vectors} to complement node attributes using structural features based on the graph (in this case, netlist). These embedding vectors are informative for predicting properties, i.e., the labels. Embedding learning  has achieved great success in the field of Natural Language Processing (NLP), where methods such as Word2Vec \cite{mikolov2013distributed} and GloVe \cite{pennington2014glove}, allow us to learn meaningful vector representations of words and use them for various downstream tasks. Crucially, the CongestionNet uses informative cell attributes (cell size and pin count) alone as the input to the GAT and does not use any embedding encoding the netlist structure. \par

In this work, we show that it is possible to create high quality structural embeddings, based on matrix factorization techniques, to enhance the node feature quality. We show that it significantly improves GNN-based congestion prediction performance. Our key findings are :

\begin{itemize}
    \item The three most popular and mainstream embedding methods for node-level embedding learning -- Node2vec \cite{grover2016node2vec}, LINE \cite{tang2015line}, and DeepWalk \cite{perozzi2014deepwalk} -- require post-processing (alignment) to be usable for cross-graph prediction.
    \item Matrix-factorization based embedding learning \cite{qiu2018network,chanpuriya2020infinitewalk} (a less popular method) combined with subgraph level training is faster, more effective, and can generalize to unseen netlists.
    \item Concatenating cell structural embeddings with cell attributes directly improves performance. When informative cell attributes are not present, a meaningful prediction can be made from netlist embeddings alone.
    \item Instead of deep, wide GATs, wide and shallow SAGE \cite{hamilton2017} GNNs achieve superior performance. 
\end{itemize}

We present our results on the publicly available DAC 2012\footnote{\href{http://archive.sigda.org/dac2012/contest/dac2012_contest.html}{http://archive.sigda.org/dac2012/contest/dac2012\_contest.html}} benchmarks using the superblue circuit line, with additional results on the OPENROAD framework.  \par


\section{Background - EDA and similarity on graphs}

\subsection{Congestion prediction in EDA}

\begin{figure}
    \centering
    \includegraphics[width=\columnwidth, trim={1cm 15.5cm 21cm 1cm}, clip]{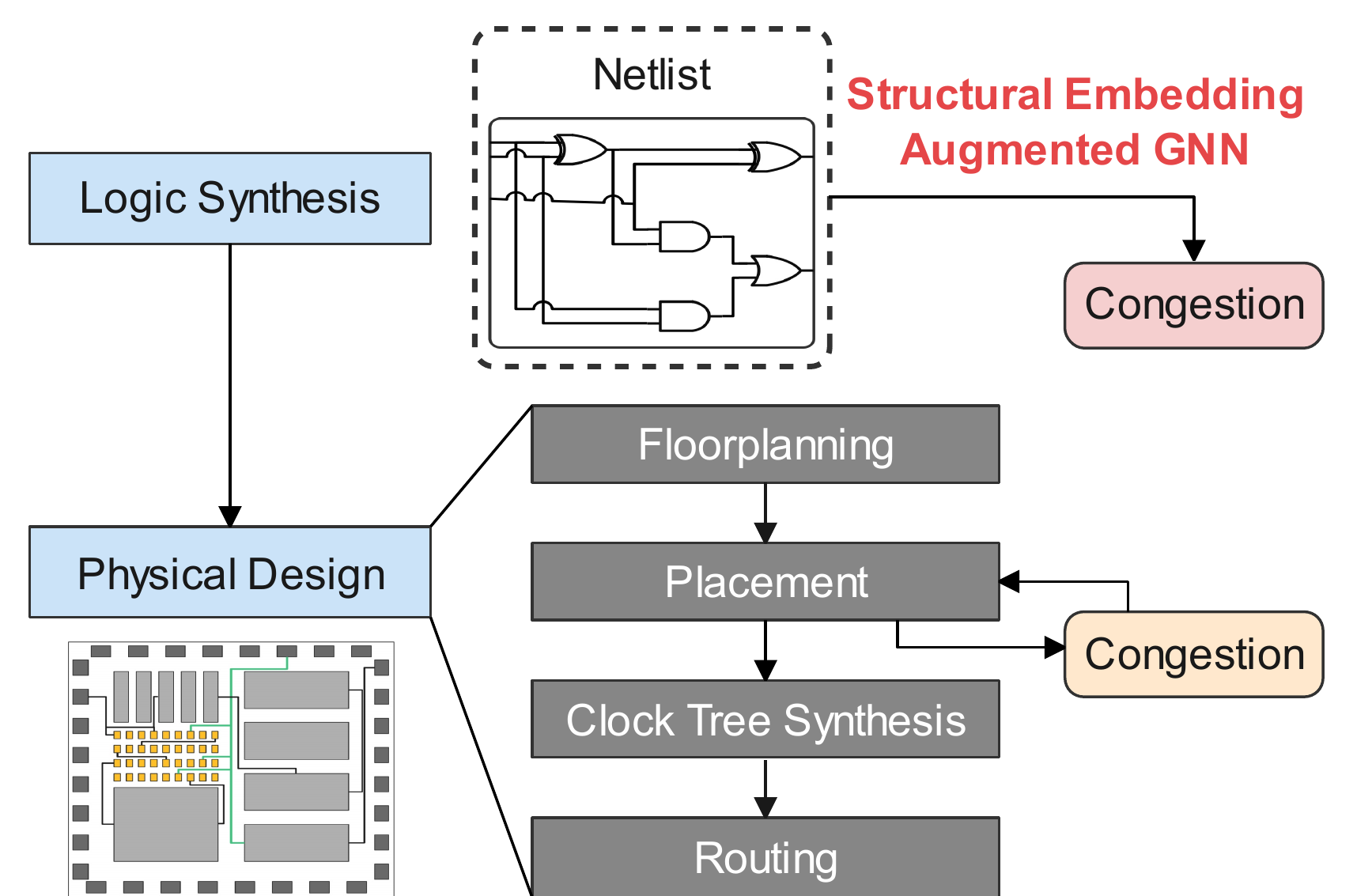}
    \caption{Congestion prediction at different EDA stages. Routing congestion is commonly estimated in placement stage and used as a feedback to improve routability of placement solution. The GNN-based method proposed in this paper predicts congestion using netlist in logic synthesis phase.}
    \label{fig:edaworkflow}
\end{figure}

In the EDA workflow, Register Transfer Level (RTL) design in VDHL or Verilog is converted to a physical layout for manufacturing through logic synthesis and physical design (Figure \ref{fig:edaworkflow}). In the physical design stage, the circuit elements are placed on the circuit board and this is followed by the routing step. Although the global routing result provides a good estimation of routing congestion \cite{liu2013nctu,xu2009fastroute}, an awareness of high congestion areas at an early design stage is of great importance to provide fast feedback and shorten design cycles.  \par

Multiple works have attempted to predict detailed routing congestion in the placement step in an effort to optimize routability of the placement solution. In \cite{saeedi2006prediction}, a probabilistic routing model is used to estimate routing demand during placement. In \cite{spindler2007fast}, the authors proposed a fast congestion prediction tool named RUDY, which estimates the wire density within the enclosing rectangle of a net using HPWL. A variety of techniques can be used to improve routability when knowledge of the congestion is available. In POLAR 2.0 \cite{lin2014polar}, cells that are estimated to have high congestion are spread out and inflated to distribute routing demand more evenly. Alternatively, as proposed in \cite{li2007routability}, an appropriate amount of white space can be allocated to different areas according to the congestion estimation. All these techniques are implemented in the placement step and need the position information of cells. \par

To avoid the high computation cost of placement, it is more useful to be able to predict congestion in the logic synthesis phase. A few works have tried to identify network structures in a synthesized netlist that can indicate high routing congestion in local areas. As shown in  \cite{kudva2002metrics}, the size of the local neighborhood is a simple but effective metric that can be used to approximate congestion. Adhesion of a logic network, defined as the sum of min-cuts for all node pairs in a graph, is a good proxy of peak congestion as shown in \cite{kudva2002metrics}. In \cite{jindal2010detecting}, the authors introduced the Groups of Tangled Logic (GTL) metric based on Rent's rule. This metric is able to detect clusters of cells that have the potential for high congestion. \par

A Graph Neural Network (GNN)-based method, named CongestionNet, has been proposed recently in~\cite{kirby2019congestionnet} to predict congestion based on a synthesized netlist. As mentioned previously, Kirby et al.\ trained a GAT-based model using cell features as input to predict the final routing congestion, without any embedding to encode the structural properties of the netlist. The key difference between our approach and their model lies in our construction of an embedding pipeline for EDA netlists. We begin by discussing similarity notions in the context of embedding learning on graphs. \par


\subsection{Proximity vs structural node similarity}
\begin{figure}
    \centering
    \includegraphics[trim={1.5cm, 20cm, 25cm, 1cm}, clip, width=0.8\columnwidth]{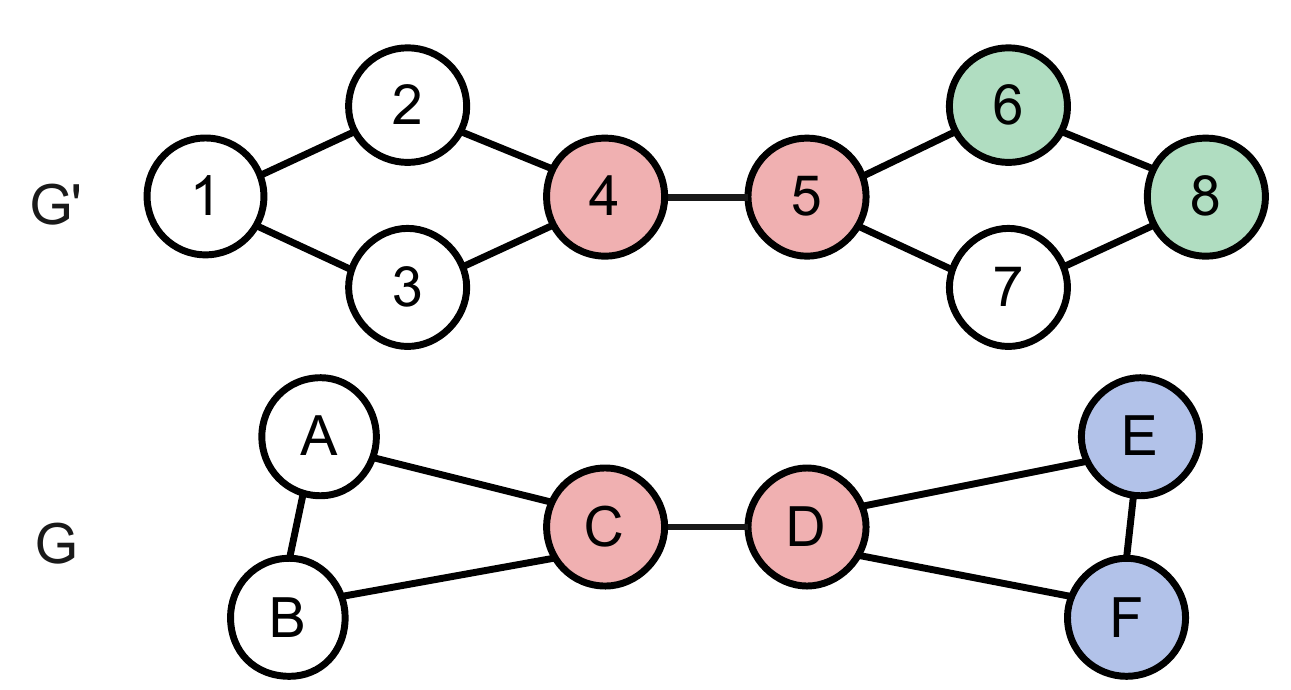}
    \caption{Structural (red) vs proximity-based (blue and green) similarity across two disjoint graphs. In particular, $4,5$ and C,D are similar even though they belong to two different graphs.}
    \label{fig:structvsprox}
\end{figure}

The most straightforward notion of similarity between two vertices $i,j$ in a graph is their \textbf{proximity} in the graph. Neighboring vertices are the most similar, and the similarity decreases as more edges must be traversed to travel from one vertex to the other. Separately from this, we may consider the notion of \textbf{structural} similarity, which relates to properties of a node such as its degree, spectral properties, etc. Two nodes can be structurally similar even if they belong to two different graphs.  The contrast between these two ideas of similarity is depicted in Figure~\ref{fig:structvsprox}, where we outline structural similarity. We now move to a detailed discussion of these similarity metrics and their suitability for EDA. \par

\section{Background - EDA, embedding and GNNs}

\subsection{From proximity measures to embeddings}

The process of node embedding involves learning a free vector $\boldsymbol{e}_v$ for each node. If $i,j$ are nodes sharing some chosen idea of similarity -- either proximity or structure-based -- $\boldsymbol{e}_i,\boldsymbol{e}_j$ should be encoded similarly in the latent space, with the similarity usually measured by the dot product $\boldsymbol{e}_i\cdot\boldsymbol{e}_j$. Methods that encode proximity similarity include random-walk based embedding methods like Node2vec~\cite{grover2016node2vec}, LINE~\cite{tang2015line}, and DeepWalk~\cite{perozzi2014deepwalk}. Methods that instead utilize structural similarity to learn embeddings include GraphWAVE \cite{donnat2018learning}, Role2vec \cite{ahmed2018learning}, and struct2vec \cite{ribeiro2017struc2vec}. \par

\subsection{Random-walk based embedding method}

Random-walk based embedding methods like Node2vec, LINE and DeepWalk are widely used in network embedding. These methods are derived from the skip-gram encoding method Word2vec~\cite{mikolov2013_word2vec} in Natural Language Processing (NLP). The methods encode a proximity relationship, and neighbouring nodes have similar embeddings using these methods. Despite their effectiveness in NLP embedding and network embedding tasks, there are two aspects of EDA that pose difficulties for standard random-walk based methods. First, the typical circuit is extremely large compared to standard graphs in the machine learning literature and comparable to the largest social network graphs. The netlists in industrial level design have hundreds of millions of nodes. Second, in the congestion prediction context, the desired prediction is often on the unseen cells in a new circuit. Thus, the train and test graphs are distinct. As random-walk based embedding only captures the proximity similarities of nodes within the same graph, training and testing on distinct graphs requires extra alignment post-processing~\cite{heimann2018regal,chen2020cone}, which is both challenging and extremely time consuming. \par

\subsection{Embedding alignment problem}

Given a graph $\mathcal{G}$, the $d$-dimensional output $\boldsymbol{X}$ of a node embedding algorithm optimizes some function of $\langle \boldsymbol{X_i, X_j} \rangle$. The proximities between nodes are preserved if we apply an orthogonal transformation matrix $\boldsymbol{Q}$ ($\boldsymbol{\tilde{X}} = \boldsymbol{XQ}$.) Consider two embeddings $\boldsymbol{X,X'}$ obtained from applying any proximity-based embedding method on two graphs $\mathcal{G},\mathcal{G}'$. For the sake of simplicity, we consider both to have the same number of nodes. $\boldsymbol{{X'},X}$ can differ via the proximity-preserving $\boldsymbol{Q}$.  \par

One way to formulate the graph alignment task is to identify an orthogonal $d\times d$ matrix $\boldsymbol{Q}$ so that we minimize the distance between $\boldsymbol{\tilde{X}} = \boldsymbol{XQ}$ and $\boldsymbol{X}'$. A caveat is that even if $\mathcal{G}$ and $\mathcal{G}'$ are identical, the ordering of the vertices $\mathcal{V}$ and $\mathcal{V}'$ may not be the same. We should therefore consider all possible permutations $\boldsymbol{PX'}$, where $\boldsymbol{P}$ is a permutation matrix of shape $|\mathcal{V}| \times |\mathcal{V}|$. \par

\begin{figure}
\centering
    \includegraphics[trim={1.5cm, 14.5cm, 21cm, 1.5cm}, clip, width=0.9\columnwidth]{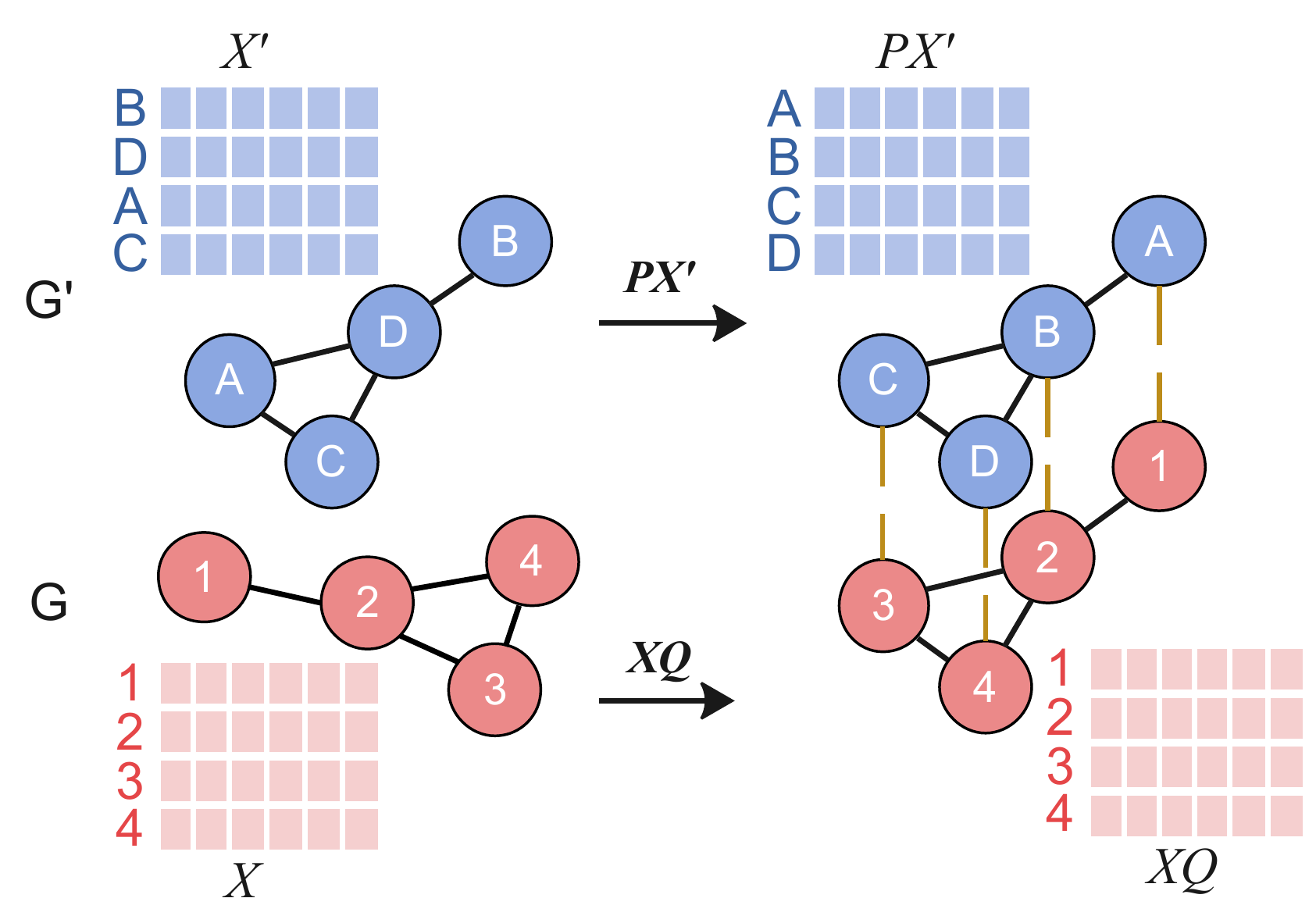}
    \caption{Wasserstein-procrustes alignment method, such as CONE-ALIGN, uses a permutation matrix and a rotation jointly to align two graphs $\mathcal{G},\mathcal{G}'$ and embeddings $\boldsymbol{X,X'}$ through $\boldsymbol{PX'}$ and $\boldsymbol{XQ = \tilde{X} }$}
    \label{fig:CONEALIGN}
\end{figure}

This formulation is \textbf{Wasserstein-Procrustes} alignment, used in network embedding alignment methods such as CONE-ALIGN~\cite{chen2020cone}, and depicted in Figure \ref{fig:CONEALIGN}.  The problem becomes to find appropriate $\boldsymbol{P,Q}$ :
\begin{equation} \underset {\boldsymbol{P} \in P_n, \boldsymbol{Q} \in O_n} {\argmin \boldsymbol{P,Q}} \; ||\boldsymbol{XQ} - \boldsymbol{PX'}||^2 
\end{equation}
Alternate alignment methods include joint factorization (REGAL~\cite{heimann2018regal}) and anchor methods (IsoRank~\cite{singh2008_IsoRank}).  \par



\subsection{Pointwise Mutual Information (PMI) Matrices}
Let $\boldsymbol{X}$, of shape $|\mathcal{V}| \times d$, be the embedding matrix representing the $d$-dimensional embeddings for all $v \in \mathcal{V}$. The similarity metric between two nodes $i,j$ is measured as $\langle \boldsymbol{X_i, X_j} \rangle$, where $\boldsymbol{X_i}$ denotes the $i$-th row of $\boldsymbol{X}$. Therefore:
\begin{itemize}
    \item The similarity pattern for all pairs $(i,j)$ of nodes is fully captured in the matrix $\boldsymbol{XX^{\top}}$, termed the \emph{PMI matrix}~\cite{Levy2014_PMI,qiu2018_NetFM}. The $(i,j)$-th entry of this matrix is equal to $\langle \boldsymbol{X_i, X_j} \rangle$.
    \item Similarity values for any pair $(i,j)$ are unchanged, and therefore the PMI matrix $\boldsymbol{XX^{\top}}$ is unchanged, if instead of $\boldsymbol{X}$, the embedding is $\boldsymbol{\tilde{X}} = \boldsymbol{XQ}$, where $\boldsymbol{Q} \in O_{d \times d}$ is orthogonal.
\end{itemize}

\par

\subsection{PMI Matrix eigendecomposition for network embedding}

The PMI matrix $\boldsymbol{XX}^\top$ under the shift $\boldsymbol{\tilde{X}} = \boldsymbol{XQ}$ becomes  $\boldsymbol{(XQ)(XQ)}^\top = \boldsymbol{XQQ}^\top \boldsymbol{X}^\top = \boldsymbol{XX}^\top$. 
From this viewpoint, if we compare graphs via the PMI matrix, there is no need to search for a suitable $\mathbf{Q}$. Based on this idea, \emph{Matrix-factorization methods} directly factor the PMI matrix -- via some unique factorization process and usually approximately via the \emph{top-k eigendecomposition} -- to obtain embeddings. We have that $\boldsymbol{XX}^\top \approx \boldsymbol{USU}^\top$, with $\boldsymbol{US^{1/2}}$ being assigned as the embedding. Uniqueness can be guaranteed by taking care of the sign of the eigenvector. With the eigendecomposition factorization strategy, it can also be shown that a permutation of the nodes does not change the similarities calculated via embeddings. In a notable analysis, NETMF~\cite{qiu2018_NetFM} unified three popular methods, Node2vec~\cite{grover2016node2vec}, LINE~\cite{tang2015line}, and DeepWalk~\cite{perozzi2014deepwalk} as implicit factorizations of a PMI matrix. In this framework, we calculate $\boldsymbol{XX}^\top$ directly, and then conduct matrix factorization via truncated eigendecomposition. Unlike $\boldsymbol{X}$ which must be estimated by random walks, $\boldsymbol{XX}^\top$ often has a closed form, and can be calculated using the adjacency matrix $\boldsymbol{A}$. Runtime is usually faster than random walk based approaches and no explicit embedding alignment is required.  \par

\subsection{Graph Neural Networks}\label{background_gnn}

We define a graph as $\mathcal{G}=(\mathcal{V,E})$ where $\mathcal{V}$ is the set of vertices, and $\mathcal{E}$ the set of edges. The adjacency matrix $\boldsymbol{A}$ has $\boldsymbol{A}_{ij} = 1$ if there is an edge $(i,j) \in \mathcal{E}$. The graphs are considered as undirected (the adjacency matrix $\boldsymbol{A}$ is symmetric) and homogeneous (all the $v \in \mathcal{V}$ are of the same type). Directed graphs and heterogeneous node sets arise in the literature and can also be solved via GNN variants, but they are not our primary focus in this work. \par

Similar to deep neural networks which repeatedly apply non-linear activation functions to form their latent representations, GNNs repeatedly apply four stages which are: neighborhood sampling, neighbor information extraction, aggregation, and representation update. The architecture is parametrized layer-wise by two separate weight matrices, $\boldsymbol{W}^{self}$ and $\boldsymbol{W}^{nbd}$. Nodes are given some initial attributes  $\boldsymbol{X}$ of shape $|\mathcal{V}| \times d$ where $d$ is the dimensionality of the initial attributes. Via multiple GNN layers, these initial attributes are propagated by the graph structure and transformed into the hidden representations of the GNN, and finally to the target labels. A sample workflow, consisting of the hidden layer iterates $\boldsymbol{h}^t_v$ for each node $v \in \mathcal{V}$ for layer depth $t = 0,\dots, T$, with the aggregate function being the mean aggregation is shown in Algorithm \ref {GNNalgo}. The schematic of a GNN predictor of our method in demonstrated in Figure~\ref {fig:gnnschematic}. \par

\begin{algorithm}
 \hspace*{\algorithmicindent} \textbf{Input :} Graph $\mathcal{G}=(\mathcal{V},\mathcal{E})$, initial input node attributes $\boldsymbol{h}^0_v$, layer depth $T$\\
 \hspace*{\algorithmicindent} \textbf{Output :} A final set of GNN node representations after T GNN layers $\boldsymbol{h}^T_v$, $\forall v \in \mathcal{V}$ 
\begin{algorithmic}[1]
\For{$i\gets1,2, \dots T$}
\For{$v \in \mathcal{V}$}
\State AGGREGATE : $\boldsymbol{m^i_v} = \sum_{u \in \mathcal{N}(v)} \frac {\boldsymbol{h^{i-1}_u} } {|\mathcal{N}(v)|}$ 
\EndFor
\For{$v \in \mathcal{V}$}
\State UPDATE : $\boldsymbol{h^i_v} = \sigma (\boldsymbol{W^{self}_{v} h^{i-1}_v + W^{nbd}_{v} m^i_v}  )$  
\EndFor
\EndFor
\end{algorithmic}
 \caption{Training loop for a GNN using degree, embedding and node attribute features}
\label{GNNalgo}
\end{algorithm}

\section{Dataset generation for congestion prediction}

As a first step, we frame the congestion prediction problem as a problem solvable by GNNs --- the \emph{node regression problem}, where continuous-valued labels are provided on some nodes (the training set) and GNNs predict this label on other nodes (the test set) with a held-out node set (validation set).


\begin{itemize}
    \item We extract two publicly available netlist sets: Superblue circuit line from DAC 2012 \cite{viswanathan2012dac} which we place via DREAMPLACE \cite{lin2020dreamplace} as well as a collection of circuits provided with the OPENROAD framework \cite{ajayi2019openroad}. We convert them into graphs. (Details in Tables \ref{dsetsb} and \ref{dsetopenroad}.)
    \item During the global routing phase, we save the position of a cell during an iteration along with the $2$-D congestion label and convert the grid congestion value to cell label.
    \item The dataset is then divided into test and train graphs. We train a GNN to create accurate predictions (in correlation terms) for the test graphs' congestion.
\end{itemize}

 \par

\begin{table}[]
\centering
\caption{Details of the superblue dataset (DAC 2012)}
\begin{tabular}{|l|l|l|l|}
\hline
\textbf{Circuit name} & \textbf{Nodes}   & \textbf{Terminals} & \textbf{Nets}    \\ \hline
\multicolumn{4}{|c|}{Train set} \\
\hline
Superblue2   & 1014029 & 92756     & 990899  \\ \hline
Superblue3   & 919911  & 86541     & 898001  \\ \hline
Superblue6   & 1014209 & 95116     & 1006629 \\ \hline
Superblue7   & 1364958 & 93071     & 1340418 \\ \hline
Superblue9   & 846678  & 57614     & 833808  \\ \hline
Superblue11  & 954686  & 94915     & 935731  \\ \hline
Superblue14  & 634555  & 66715     & 619815  \\ \hline
\multicolumn{4}{|c|}{Train graph for ablation study - also in normal train set} \\
\hline
Superblue16  & 698741  & 18291     & 697458  \\
\hline
\multicolumn{4}{|c|}{Validation set} \\
\hline
Superblue12  & 1293433 & 15349     & 1293436 \\ \hline
\multicolumn{4}{|c|}{Test set} \\
\hline
Superblue19  & 522775  & 16678     & 511685  \\ \hline

\end{tabular}
\label{dsetsb}
\end{table}

\begin{table}[]
\centering
\caption{Details of the OPENROAD dataset}
\begin{tabular}{|l|l|l|l|}
\hline
\textbf{Circuit name} & \textbf{Nodes} & \textbf{Terminals} & \textbf{Nets}    \\ \hline
\multicolumn{4}{|c|}{{Train set}} \\
\hline
gcd         & 343   & 54    & 414  \\ \hline
ibex        & 22279 & 264   & 24368 \\ \hline
aes         & 23669 & 388   & 24458 \\ \hline
tinyRocket  & 33225 & 269   & 37250 \\ \hline
jpeg        & 89737 & 47    & 94139 \\ \hline
bp\_multi   & 93528 & 1453  & 110847 \\ \hline
Dynamic\_node   & 12112  & 693    & 14736 \\ \hline
bp\_fe      & 32453 & 2511  & 38672 \\ \hline
\multicolumn{4}{|c|}{Train graph for ablation study - also in normal train set} \\
\hline
bp & 151415 & 24 & 179901 \\ \hline
\multicolumn{4}{|c|}{Validation set} \\
\hline
bp\_be      & 54591 & 3029  & 64829 \\ 
\hline
\multicolumn{4}{|c|}{Test set} \\ \hline
swerv & 102034 & 2039 & 114079 \\ \hline

\end{tabular}
\label{dsetopenroad}
\end{table}

\begin{figure*}
\centering
    \includegraphics[trim={1cm, 12cm, 8cm, 1cm}, clip, width=0.8\textwidth]{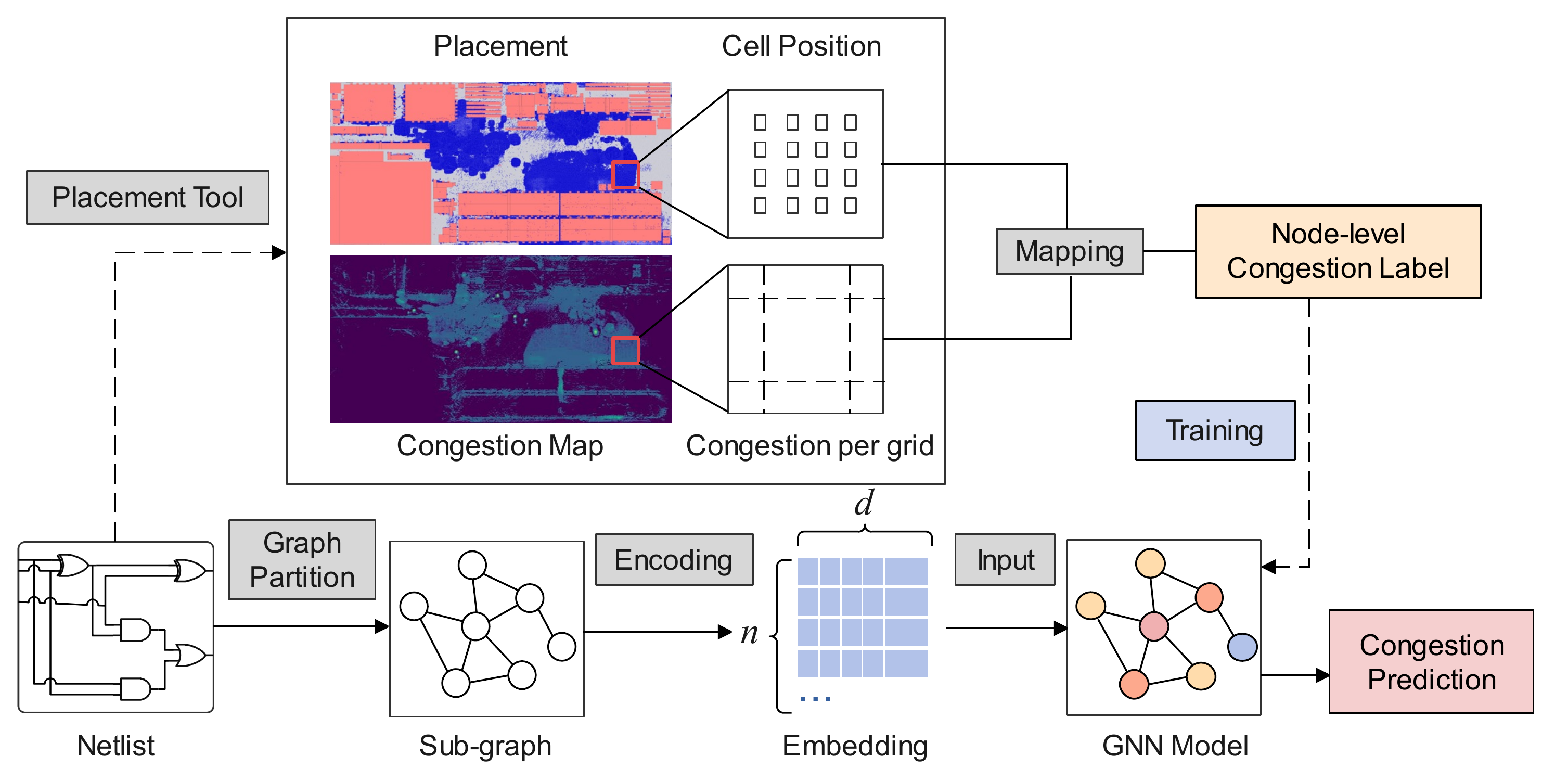}
    \caption{The training (dashed arrows) and inference (solid arrows) flow of our GNN framework.}
    \label{fig:congflow}
\end{figure*}


The synthesized design is represented by a netlist of circuit elements. Every circuit element contained in the netlist is regarded as a node in the graph. The edges represent interconnections between circuit elements as defined in the nets of the netlist. We assume the set of nodes in each net is fully connected. Macros and terminals are removed from the graph, because they are normally manually placed in practice and their high node degrees compared to standard cells reduce training efficiency. This leaves only cells as circuit elements. Nets with degree more than $10$ are excluded from the final graph as they introduce cliques too large to work with efficiently. The informative cell features (pin number, cell size) are stored as node features and can be used in combination with structural embeddings as GNN inputs. This follows the flow of CongestionNet \cite{kirby2019congestionnet}.   \par

The workflow to generate labels for training is shown in Figure \ref{fig:congflow}. For superblue circuit lines, we use DREAMPLACE for placement and NCTUGR 2.0 for congestion map generation, while for OPENROAD datasets, we use the built-in placer RePlAce for placement and FastRoute for congestion map generation \cite{dai2011nctu,liu2013nctu, xu2009fastroute}.   \par
The board is partitioned into grids of size $C_x, C_y$, with the congestion value for each grid cell computed as the wiring demand divided by the routing capacity. The output along the $z$-axis is reduced by a max function, with the congestion of layer $0$ removed due to lack of routing capacity. The position of every cell $(x,y)$ with grid spacing $C_x,C_y$ can be converted into $(\lfloor \frac{x}{C_x} \rfloor, \lfloor \frac{y}{C_y} \rfloor)$, and we set the label of the cell as the congestion at $(\lfloor \frac{x}{C_x} \rfloor, \lfloor \frac{y}{C_y} \rfloor)$. All labels are normalized to a fixed range of $[-6,6]$. Our focus is on predicting congestion due to local logic structure, which manifests itself on lower metal layers. Therefore, we use congestion labels from the lower half of the metal layers to train and evaluate the model. Results using overall congestion are also presented for comparison. We now describe the partitioning and embedding steps; it is in these steps where our approach differs most significantly from CongestionNet.  \par

\section{Embedding-enhanced GNN training framework}

\begin{figure*}
    \centering
    \includegraphics[trim={1cm, 1.5cm, 1.5cm, 2cm}, clip, width=0.85\textwidth]{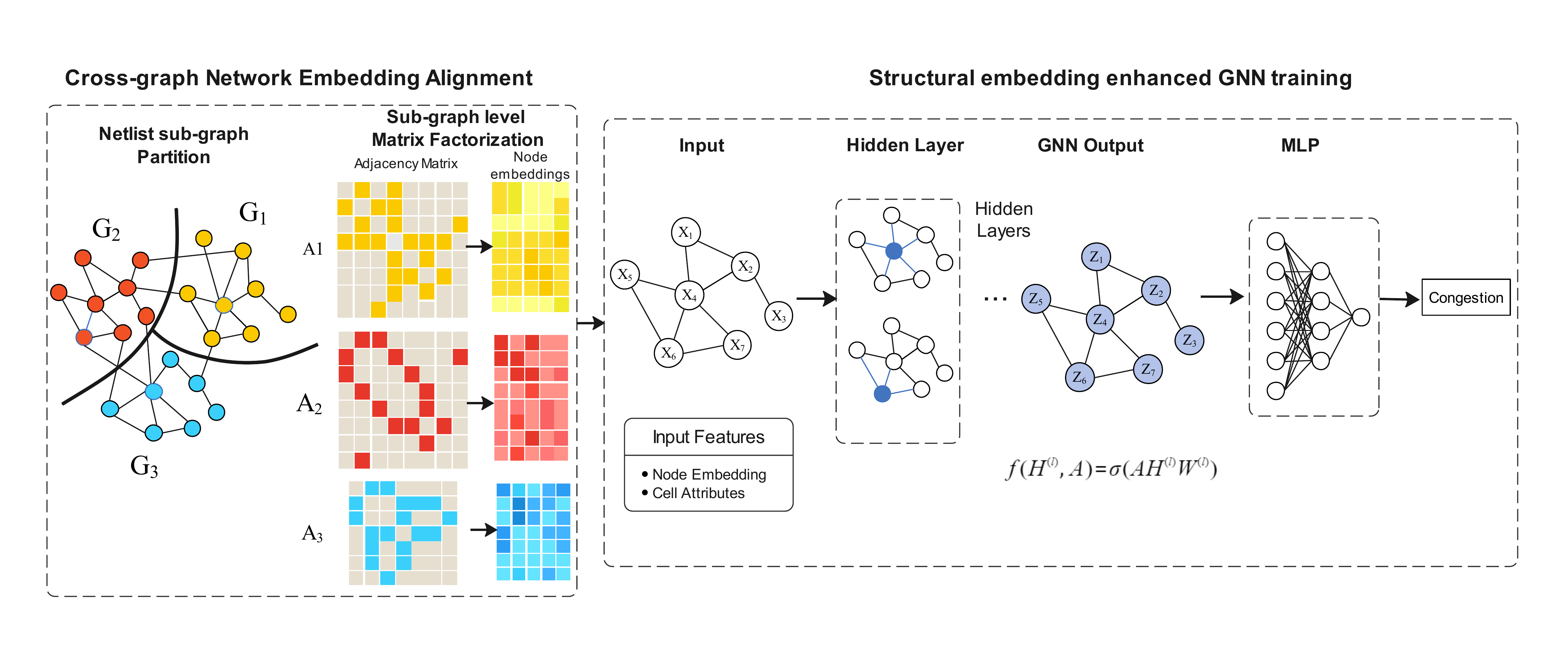}
    \caption{Our system design for structural embedding-enhanced GNN with a MLP post predictor for congestion prediction}
    \label{fig:gnnschematic}
\end{figure*}

The overview of our proposed framework is shown in Figure \ref{fig:gnnschematic}. We now sequentially describe all the sub-processes for training and inference of a complete GNN model.

\subsection{Sub-graph partition and batch-training}

The graph from the netlist of a complete circuit is too large for direct matrix factorization and must be partitioned into clusters. We use the METIS partitioning tool used in the ClusterGCN \cite{chiang2019cluster} framework to partition graphs for training. In METIS, the graph is coarsened by repeatedly merging nodes that are connected by highly weighted edges. After coarsening, the graph is bisected and resolved to its original position iteratively. The bisection is expanded to obtain a K-way partition. We set the clustering configuration to yield clusters of $\approx 5000$ nodes each, with all superblue circuits being partitioned $150$ times and OPENROAD circuits set to the closest integer to reach the target cluster size. The combination of matrix factorization with clustering in a GNN is, to our knowledge, novel even outside the EDA context. \par

\subsection{Matrix Factorization for embedding generation}

Given a partition $P$ with associated adjacency, degree and Laplacian matrices $\boldsymbol{A_P, D_P, L_P}$, we use a modified version of the InfiniteWalk \cite{chanpuriya2020infinitewalk} matrix factorization method to form the embedding, with $\boldsymbol{11}^{\top}$ the all-ones matrix, and $L,T > 0,H \geq L$ being hyperparameters:

\begin{itemize}
    \item $\boldsymbol{M'_P}  = \boldsymbol{11}^{\top} + {Tr\boldsymbol{(D_P) D^{-1/2}_P L_P D^{-1/2}_P}} $. 
    \item $\boldsymbol{M''_P} = \boldsymbol{11}^{\top} + {\frac{\boldsymbol{M'_P}}{T}}$. 
    \item Clamp $\boldsymbol{M''_P}$ to range $[L,H], L > 0$. 
    \item $\boldsymbol{M''_P} \leftarrow \log (\boldsymbol{M''_P})$ (entrywise log). 
\end{itemize}

The matrix $\boldsymbol{M''_P}$ is the PMI matrix, and a truncated eigendecomposition yields the embedding. The hyperparameter choice depends on the graph structure. While $L,H$ control the numerical stability, $T$ controls the extent to which a node may influence its neighbour. High values of $T$ generally lead to similar embeddings graph-wide, an undesirable outcome termed \textbf{oversmoothing}, which harms performance. 

The factorization generates a K-dimensional embedding vector for every cell. This is run for every partition generated by the METIS step and the embeddings are cached. The embeddings are comparable between partitions due to the usage of PMI matrices and require no extra alignment. 

For use with node attributes $\boldsymbol{X}$, the embeddings $\boldsymbol{E}$ and attributes are concatenated. These are fed into a special GNN architecture called SAGE~\cite{hamilton2017}. For the detailed formulation of SAGE please refer to Section~\ref{background_gnn}. To further refine the final representation with information passed directly from the original node features, the embedding and the attribute are concatenated with the final layer of SAGE output and fed into an MLP to generate the final congestion estimate $\hat{Y}$:  \par

$$ \hat{Y} = \mathrm{MLP} \Big(\big[[\boldsymbol{X};\boldsymbol{E}] ; \mathrm{SAGE}([\boldsymbol{X}; \boldsymbol{E}])\big]\Big)\,, $$ 
where $[\,\,;\,]$ represents concatenation.




\section{Training and inference details and results}

\subsection{GNN architecture details}

Graph neural networks are prone to oversmoothing, where the predictions for all nodes become very similar, and this arises more often in deeper GNNs. Accordingly, as opposed to CongestionNet which uses $16$ hidden states and a depth $8$ GAT~\cite{kirby2019congestionnet, velickovic2018}, we use SAGE~\cite{hamilton2017} (SAmple and aGgregatE) convolutions to construct our GNNs and set them to have $2$ hidden layers of size $200$ and $160$. The input to the GNN is the node embedding concatenated with the attributes. The output of the last hidden layer is concatenated with the input and fed to an MLP with two hidden layers, each of size $150$, to obtain the final congestion predictions. Squared error loss is used to train with ADAM \cite{kingma2014adam} as the optimizer, with each minibatch containing one cluster. The parameters $L,H$ are set to $10^{-10}, 10^6$ respectively. The dimension of the embedding is set to $4$.   \par

\begin{table*}[t]
\centering
\caption{Results obtained on all superblue circuits (DAC 2012)}
\begin{tabular}{|l|l|l|l|l|l|l|l|l|l|l|l|l|l|}
\hline
\multicolumn{13}{|c|}{Congestion prediction study correlation metrics}\\
\hline
\multicolumn{1}{|c|}{ } & \multicolumn{6}{|c|}{Lower level congestion} &
\multicolumn{6}{|c|}{Overall congestion}\\
\hline
Methods              &  \multicolumn{2}{|c|}{Pearson} &  \multicolumn{2}{|c|}{Spearman} & \multicolumn{2}{|c|}{Kendall} & \multicolumn{2}{|c|}{Pearson} & \multicolumn{2}{|c|}{Spearman} & \multicolumn{2}{|c|}{Kendall} \\ \hline
           &    Node     &      Grid     &      Node     &    Grid     &  Node        &  Grid   &    Node      &     Grid     &     Node     &    Grid     &     Node      & Grid  \\ \hline
Adhesion metric     &   0.09       &   0.16        &   0.06       &     0.20    &   0.06       &  0.14  &    0.08     &     0.16      &       0.07     &   0.20      &    0.05       &  0.15  \\ \hline
Neighbourhood metric  &    0.02     &    0.04       &      0.18   &   0.27       &     0.13     & 0.18 &   0.03      &    0.08      &     0.18     &    0.27      &    0.13       &  0.18 \\ \hline
GTL metric         &     0.02    &  0.01         &    0.14      &      0.23     &   0.10         &  0.16 &       0.01     &      0.05      &       0.14   &    0.24      &  0.10        &  0.16 \\ \hline
CongestionNet   &    0.26     &     0.35     &   0.27       &     0.33    &    0.19      &  0.24 & 0.19         &     0.27     &  0.24        &   0.26      &    0.17       & 0.22 \\ \hline
Embedding-enhanced GNN (ours)          &  0.31     &    0.43   &   0.34  &   0.44   &      0.25      &      0.31  &    0.24    &    0.33  &    0.26      &    0.38  &   0.20     &    0.27   \\ \hline

\multicolumn{13}{|c|}{Embedding ablation findings - train set reduced to one graph}\\
\hline
GNN (no embedding)   &    0.27     &   0.36       &     0.30    &     0.40    &     0.19     &    0.27 &    0.21     &   0.28        &  0.22       &    0.32     &  0.15        &  0.23 \\ \hline
Embedding-enhanced GNN (ours)    &    0.30     &  0.41        &  0.32        &      0.42   &     0.24     &   0.30  &   0.22      &     0.30     &  0.24       &    0.35     &    0.18      &  0.26  \\ \hline

\multicolumn{13}{|c|}{Comparing embeddings only}\\
\hline
LINE (aligned)       &  0.10       &    0.14      &     0.15    &  0.21       &     0.09     & 0.15   &  0.06       &   0.09       &    0.12     &  0.22       &    0.09      & 0.16     \\ \hline
LINE (unaligned)     &  0.02        &   0.06       &   0.03               &    0.06      &   0.02   & 0.04         &      0.03    &    0.06     &  0.04       &     0.05     & 0.01  & 0.03 \\ \hline
Node2Vec(aligned)    &   0.10      & 0.15         &   0.10      &   0.14      &    0.07      & 0.10   &     0.09    &  0.13        &    0.10     &  0.12       &    0.08      & 0.10    \\ \hline
Node2Vec(unaligned)  &    0.03     &   0.07       &      0.02   & 0.06         &  0.01        &  0.04   &   0.05      & 0.07          &  0.04       & 0.05         &    0.03      & 0.04     \\ \hline
DeepWalk(aligned)    &   0.09      &    0.15      &     0.12    & 0.19         &    0.06      & 0.11    &     0.08    &  0.12        &    0.10     & 0.13         &  0.09        & 0.12     \\ \hline
DeepWalk(unaligned)  &   0.00      &    0.04      &    0.02     &   0.04      &     0.02     & 0.03    &     0.02    & 0.02          &    0.03  & 0.02         &   0.02       & 0.01     \\ \hline
GNN(embedding only)  &  0.16       &    0.25      &  0.17       &   0.26      &   0.15       &  0.20  &    0.14     &     0.17     &   0.16      &    0.25     &     0.13     & 0.17  \\ \hline

\end{tabular}
\label{superblue}
\end{table*}

\begin{table*}[t]
\centering
\caption{Results obtained on all OPENROAD circuits}
\begin{tabular}{|l|l|l|l|l|l|l|l|l|l|l|l|l|l|}
\hline
\multicolumn{13}{|c|}{Congestion prediction study correlation metrics}\\
\hline
\multicolumn{1}{|c|}{ } & \multicolumn{6}{|c|}{Lower level congestion} &
\multicolumn{6}{|c|}{Overall congestion}\\
\hline
Methods              &  \multicolumn{2}{|c|}{Pearson} &  \multicolumn{2}{|c|}{Spearman} & \multicolumn{2}{|c|}{Kendall} & \multicolumn{2}{|c|}{Pearson} & \multicolumn{2}{|c|}{Spearman} & \multicolumn{2}{|c|}{Kendall} \\ \hline
           &    Node     &      Grid     &      Node     &    Grid     &  Node        &  Grid   &    Node      &     Grid     &     Node     &    Grid     &     Node      & Grid  \\ \hline
Adhesion metric     &  0.12       &  0.13        &   0.09      &   0.14      &     0.07     & 0.10  &     0.10    & 0.12          &   0.06      &   0.12      &     0.05     & 0.09   \\ \hline
Neighbourhood metric  &   0.13      &    0.12      &   0.08      &    0.13      &    0.10      & 0.06 &    0.12     & 0.10          &     0.06    &  0.13       &   0.05       & 0.10  \\ \hline
GTL metric         &  0.10       &     0.12     &     0.08    &      0.14   &   0.06       & 0.10   & 0.09        &     0.10     &   0.06      &    0.13     & 0.05         & 0.10  \\ \hline
CongestionNet   &     0.20    & 0.23          &   0.16      & 0.16        &    0.13      & 0.11  &            0.17      & 0.19         &    0.15     & 0.14   &   0.12    & 0.10  \\ \hline
Embedding-enhanced GNN (ours)      &  0.23       &     0.26     &   0.20      & 0.19        &     0.18     & 0.17  &    0.19     & 0.22          &      0.16   & 0.16         &   0.13       & 0.12   \\ \hline
\multicolumn{13}{|c|}{Embedding ablation findings - train set reduced to one graph}\\
\hline
GNN (no embedding)   &   0.18      &  0.20        &   0.15      & 0.14        &   0.12       & 0.10  &     0.13    & 0.15          &   0.11      &  0.09        &    0.08      & 0.10   \\ \hline
Embedding-enhanced GNN (ours)    &  0.20       &       0.23   &     0.17    & 0.17         &    0.13      & 0.11  &     0.14    &   0.18       &    0.12     & 0.14        &  0.09        & 0.10  \\ \hline
\multicolumn{13}{|c|}{Comparing embeddings only}\\
\hline
LINE (aligned)      &    0.10     & 0.11          &    0.09     &  0.07        &  0.07        & 0.05  &     0.10    &    0.11      &    0.09     &  0.08       &    0.07      & 0.06  \\ \hline
LINE (unaligned)     &   0.00      & 0.00         &    0.00     &  0.00       &    0.00      & 0.00 &      0.01   &      0.01    &    0.01     &  0.01       &     0.01     & 0.01 \\ \hline
Node2Vec(aligned)    &  0.11       &  0.12        &    0.09     &  0.07       &       0.07   & 0.05  &     0.12    & 0.12          &    0.10     &    0.07     &     0.08     &  0.05  \\ \hline
Node2Vec(unaligned)  &   0.01      & 0.01         & 0.02         &  0.00       &     0.01     & 0.00  &    0.02     & 0.03         & 0.03         & 0.01         &     0.02     &   0.01  \\ \hline
DeepWalk(aligned)    &  0.10       &    0.11      &    0.08     &   0.06      &    0.06      &  0.04 &    0.12      &      0.13    &  0.10       &    0.07     &    0.08      &  0.05   \\ \hline
DeepWalk(unaligned)  &  0.02       &  0.01        &   0.01      &    0.02     &     0.01     & 0.02   &   0.01      &   0.01       &   0.01      &    0.02     &     0.01     & 0.02\\ \hline
GNN(embedding only)  &   0.13      & 0.15          &     0.11    & 0.15         &   0.09       & 0.11  &      0.12   & 0.13          &   0.10      &  0.14       &   0.08       & 0.10 \\ \hline

\end{tabular}
\label{openroad}
\end{table*}

\subsection{Training and inference}

During training, the loss is optimized over the nodes of the graph, and accuracy on the nodes defines our {graph level} metrics. During inference, an unseen graph is provided and the embedding is computed. We do not need the additional alignment step required for using proximity based embedding here. Once the new graph's embedding is computed, we concatenate it with the new graph's attributes (if available) and output a graph level prediction. The unseen graph has a ground truth position provided in the evaluation phase. For evaluation, along with graph level we consider a \textbf{reconstructed grid}, which consists only of the grid cells with at least one non-terminal node, and is reconstructed by averaging the labels and predictions of nodes in a grid cell.   \par

In our experiments, we consistently found that :

\begin{itemize}
    \item The usage of average labels, where each cell instead inherits the congestion in $(\lfloor \frac{x}{C_x} \rfloor, \lfloor \frac{y}{C_y} \rfloor)$ divided by the total number of cells in $(\lfloor \frac{x}{C_x} \rfloor, \lfloor \frac{y}{C_y} \rfloor)$, is not superior in terms of final performance, although it does help in terms of rapidly converging to the optimum.
    \item The usage of pin features, beyond simply counting the pins of each cell, does not help the outcome, and usually prevents good fitting of models.
    \item The addition of terminals back to the graph also hinders training of GNNs. This may be understood to arise from the size of the macros causing problems in label creation for the graph.
\end{itemize}

\subsection{Correlation metrics for evaluation}
We use the following three metrics of correlation to measure performance, evaluated both on graph and grid level.  \par

\begin{itemize}
    \item Pearson correlation coefficient (PCC). Given two random variables $X,Y$, the PCC between $X,Y$ is defined as $E[\Tilde{X}\Tilde{Y}]$, where $\Tilde{X} = \frac{X - E[X]}{\sqrt{Var(X)}}$ and similarly for $Y$.
    \item Spearman correlation, which calculates PCC but replaces $X,Y$ with their rank among all $X,Y$ in observed sample.
    \item Kendall correlation : Let $N$ samples be present of $X,Y$. Two pairs $(X_i,X_j)$ and $(Y_i,Y_j)$ are concordant if $(X_i - X_j)(Y_i - Y_j) > 0$. That is, if $X_i > X_j$ then $Y_i > Y_j$ and vice versa. Kendall correlation is the difference between concordant and non-cordant pairs. divided by the total number of pairs ${N}\choose{2}$. This is the only metric used in~\cite{kirby2019congestionnet}. We add the other two metrics for more complete evaluation as especially Pearson can capture raw values which Spearman and Kendall cannot.
\end{itemize}

Before evaluation, both the prediction and the label have some (very low) noise added to them. This is to break ties randomly, as the demand values are discrete and in this situation the concordance of pairs is not meaningful.  \par

\subsection{Learning-free structure-based benchmarks}

We use as benchmarks three methods which are also found in~\cite{kirby2019congestionnet}. In each, a value is computed per cell and the corresponding correlation metrics computed using this value as a congestion prediction. In case of cross-validation of parameters, lower-level grid Kendall correlation coefficient is used to order settings of computation.  \par
\begin{itemize}
    \item Neighborhood size: The $k$-th neighborhood size of a node $v$ is the number of nodes, reachable from $v$, within geodesic distance $k$. $k$ is varied from $1$ to $5$ and the best performing value is taken.
    \item GTL (Graph Tangled Logic) : This refers to a measure that examines the structure of the graph cut around node $v$. Parametrized by a term called the \textbf{rent exponent}, it seeks to adjust for the effect of cluster size. We cross-validate and report only the result with the best rent exponent.
    \item Adhesion : We grow a local neighbourhood around a node $v$ and compute all min-cuts between the node and other nodes in the neighbourhood, and take the maximum min-cut as indicative of connectivity around $v$. The neighbourhood size is varied between up to $1$ and $5$ nodes away and the best value is reported.
\end{itemize}

\begin{table*}[t]
\centering
\caption{Runtime comparison (seconds) between subgraph-level training and block sampling}
\begin{tabular}{|l|l|l|l|l|}
\hline
\multicolumn{5}{|c|}{Architecture runtime comparisons (per graph per epoch)}  \\ \hline
 & \multicolumn{2}{|c|}{Superblue} & \multicolumn{2}{|c|}{OPENROAD} \\ \hline
GNN architecture & Training time & Inference time & Training time & Inference time \\ \hline
No partitioning (ours + block sampler) & 56.2 & 65.4 & 9.8 &  14.1\\ \hline
\multicolumn{5}{|c|}{With partitioning}  \\ \hline
Our architecture & 2.2 & 6.1 & 0.22 & 2.5 \\ \hline 
CongestionNet & 6.4 & 8.7 & 0.78 & 3.8 \\ \hline
\end{tabular}
\label{runtime}
\end{table*}
\begin{table*}[t]
\centering
\caption{Matrix factorization runtime vs other embedding methods in seconds}
\begin{tabular}{|l|l|l|l|l|}
\hline
\multicolumn{5}{|c|}{Embedding runtime comparisons} \\ \hline
Embedding method & Train time & Alignment time & Train time & Alignment time \\ \hline 
Node2vec & 250.6 & 1750.4  & 45.2  & 733.5  \\ \hline
LINE     & 167.8 & 1355.2  & 19.7  & 802.1  \\ \hline
DeepWalk & 143.7 & 1566.5 & 22.1  & 783.4  \\ \hline
Ours     & 80.4 & -  & 18.2  & -  \\ \hline
\end{tabular}
\label{embruntime}
\end{table*}

\subsection{Key results on Superblue and OPENROAD }

The aim of our results is to compare with the un-augmented graph convolutional network of the same structure, as well as to compare with the previously proposed CongestionNet~\cite{kirby2019congestionnet}. In comparison with these benchmarks, our augmented network has the best performance. Results on Superblue and OPENROAD are respectively provided in tables \ref{superblue} and \ref{openroad}. All results are reported using a Tesla V100 GPU and run on a pytorch \cite{paszke2019pytorch} + Deep Graph Library (DGL) \cite{wang2019deep} codebase.   \par

We also provide comparison of the InfiniteWalk embedding as an input to the GCN versus embeddings from Node2vec, LINE, and DeepWalk, all aligned using the CONE-ALIGN method. The InfiniteWalk embedding is shown to provide superior results. We provide an unaligned case as a simple comparison to highlight the need for alignment. For multi-graph alignment, the choice of anchor graph may not be clear. As such, the training and alignment for the embedding-only case is done with exactly one training graph --- superblue$16$ and bp for DREAMPLACE and OPENROAD respectively. The results are thus a lower bound on the performance achievable with InfiniteWalk-like embedding alone. Each embedding dimension is set to $32$. InfiniteWalk-like embedding generally achieves the best performance even over the aligned proximity embeddings, as well as improving attribute-equipped GNNs. The improvement on top of attributes ranges from $10$ to $20$ \%. Relative to aligned embeddings, InfiniteWalk can be up to $60 \%$ more effective outperforms non-learning benchmarks.  \par

\subsection{Runtime comparison}

Using clustering and partitioning the graph leads to significant improvements in runtime when compared to conventional samplers. These comparisons are made in Tables \ref{runtime} and \ref{embruntime}. The without-partition version is constructed using a Multi-layer block sampler. In particular, note that in terms of embedding comparisons, our InfiniteWalk-derived embedding is far faster due to requiring no alignment time. This is actually an \textbf{underestimate} of the alignment time, due to the numerical instability of the optimization problem requiring extra time for tuning the correct regularizer value. Clustering by itself can cut up to $90 \%$ of epoch time or more, and avoiding the alignment step offers a similar speedup.  \par

\section{Conclusion, comparison, possible extensions}

In this work, we provide the GNN-based method for predicting congestion from a netlist that can operate without any useful attribute of cells, and solely on netlist structure. Our approach is inspired by the fact that classic methods for congestion estimation such as GTL \cite{jindal2010detecting}, which rely solely on structural properties, are able to pinpoint congested cells. This suggested that embedding learning for circuits might be able to capture the aspects that allowed these previous graph-based predictors to do well. Crucially, by adding the Pearson correlation, we also seek to measure the amount of prediction for the raw congestion value, and not just the rank.  \par

Our most important finding is that the most popular embedding learning methods, which are otherwise dominant in GNN-related works, do not perform particularly well in the EDA task. The success of matrix factorization paired with clustering is a novel finding in the EDA context, and potentially even outside of it. Comparison with previous methods, such as CongestionNet, shows that we improve on key aspects such as correlation metrics and runtime. This makes our method attractive even when there are attributes available.  \par 

There are currently no theoretical explanations for the success of matrix factorization when clustering is employed. We believe that understanding the types of graph structures which allow this kind of processing is important. From a practical perspective, it should be noted that proximity based methods constitute the majority of embedding research in the machine learning community, and alignment methods are also rapidly improving. Therefore, there are possible pairings of proximity based random walk embedding methods with alignments that might compete with matrix factorization. While we achieve superior performance with matrix factorization methods, random walk methods are superior in terms of scalability in memory, while matrix factorization methods are superior in terms of runtime scalability. There should be a smooth tradeoff between these two methods, which deserves further exploration.   \par

Recently, there has been interest in not only the prediction of congestion, but using such prediction as feedback \cite{chen2020pros} to placement algorithms \cite{liu2021global} to control the placement process. Although we have demonstrated successful prediction, it remains an open challenge to integrate our predictor for the control problem and improve metrics such as wirelength in the final (post-detailed routing and placement) design.  \par

\clearpage

\bibliographystyle{IEEEtran}  

\end{document}